\documentclass{llncs}
\usepackage{graphicx}

\usepackage[utf8]{inputenc} 
\usepackage[T1]{fontenc}    
\usepackage{hyperref}       
\usepackage{url}            
\usepackage{booktabs}       
\usepackage{amsfonts}       
\usepackage{nicefrac}       
\usepackage{microtype}      
\usepackage{xcolor}         
\usepackage{multirow}

\begin{document}
\title{ADoBo 2021:\\The futility of STILTs for the classification of lexical borrowings in Spanish}
%
%
\author{Javier de la Rosa\inst{1,2}\orcidID{0000-0002-9143-5573}}
%
%
\institute{
    National Library of Norway AI-Lab, Mo i Rana 8624, Norway \\
    \email{versae@nb.no} \and
    LINHD, UNED, Madrid, 28040, Spain \\
    \email{versae@linhd.uned.es}
}

%
\maketitle              
\begin{abstract}
The first edition of the IberLEF 2021 shared task on automatic detection of borrowings (ADoBo) focused on detecting lexical borrowings that appeared in the Spanish press and that have recently been imported into the Spanish language. In this work, we tested supplementary training on intermediate labeled-data tasks (STILTs) from part of speech (POS), named entity recognition (NER), code-switching, and language identification approaches to the classification of borrowings at the token level using existing pre-trained transformer-based language models. Our extensive experimental results suggest that STILTs do not provide any improvement over direct fine-tuning of multilingual models. However, multilingual models trained on small subsets of languages perform reasonably better than multilingual BERT but not as good as multilingual RoBERTa for the given dataset.

\end{abstract}
{\let\thefootnote\relax\footnotetext{\textit{IberLEF 2021, September 2021, Málaga, Spain.}\\Copyright \textcopyright\ 2021 for this paper by its authors. Use permitted under Creative Commons License Attribution 4.0 International (CC BY 4.0).}

\section{Introduction}

The sociopolitical and sociocultural changes speakers undergo tend to be somehow reflected in their lexicon \cite{soto2018loanwords}. The borrowing of words from a donor language to a recipient one is a common mechanism driving language change and word formation \cite{trask2009languages}. The lexical units being incorporated into the recipient language usually undergo morphological and phonological transformations as to better conform with the features of the recipient language \cite{poplack1988social}. Interestingly, regardless of the reasons as to why languages borrow words from others, there are a few common patterns that emerge in the process. For example, it seems like content words are much more frequently borrowed than function words, since it is more likely for a language to borrow nouns or verbs rather than prepositions or conjunctions. Nouns in particular might be benefiting from referential transparency and morphosyntactic freedom \cite{matras2020language}. However, before a word becomes fully assimilated into a recipient language with the proper morphological or orthographic modifications, it is not uncommon to see the adapted and unadapted versions of borrowings coexisting (e.g., \textit{whisky} and \textit{güisqui} are both correct in the Spanish orthography). Identifying what words enter a language and how they do so is critical for the understanding of the development of a language.

\section{Related work}

Given its global dominance in many domains of our daily lives, automatic approaches for detecting borrowings are mainly focused on words of English origin, that is, anglicism detection \cite{furiassi2007retrieval,chesley2010predicting,andersen2012semi,furiassi2012anglicization}.

For Spanish, a loanword identification algorithm for Argentine Spanish was proposed in \cite{serigos2017applying}, which provided the lemmatized form of the tokens, identified named entities, and preserved loan phrases. Although somewhat successful in the identification of anglicisms in the chosen news corpus, the algorithm was designed as a binary classifier in which every word was labeled as Spanish or English. This impacted negatively the number of false positives and made the system incapable of reliably identifying code-switching or adapted loans.

More recently, \cite{alvarez2020lazaro} showed that framing the problem as a token classification task for the extraction of emergent anglicisms in Spanish newswire provided very good results. The approach compared a conditional random fields (CRF) classifier built upon handcrafted features with word and character-level embeddings, to a bidirectional long short-term memory neural network with an extra conditional random fields layer on top (BiLSTM-CRF). The CRF-only method outperformed the neural network and achieved an F1 score of 87.82 on the test. Interestingly, the corpora made explicit distinctions between unadapted borrowings of English origin or other origin.

\section{Language models}

Previous methods have not yet relied on modern language models for the detection of borrowings. In this work, we used pre-trained transformer-based language models to approach the problem of borrowing detection as a token classification task. Given the prevalence of morphological and orthographic differences between emergent unadapted borrowings and assimilated words that have been part of a language for a long time, our hypothesis is that if a language model is capable of acquiring enough morphological and syntactical information should also be able to use this information to perform well on the detection of borrowings.

In order to test this hypothesis, we compared the results of fine-tuning BERT-based language models \cite{devlin18} on a corpus of Spanish news articles with and without supplementary training on intermediate labeled-data tasks (STILTs) \cite{phang2018sentence} related to the kind of language features that precisely makes identifying unadapted borrowings possible. Specifically, named entity recognition, part of speech, code-switching, and language identification. The rationale is that if a model is able to perform well in any of these tasks, it should also do well at detecting borrowings.

Supplementary training is a technique that supplements ``language model-style pre-training with further training on data-rich supervised tasks" \cite{phang2018sentence}. When using intermediate tasks such as natural language inference, performance on benchmarks like GLUE \cite{wang2018glue} improves over BERT \cite{devlin18} or ELMo \cite{peters18}. Moreover, it seems to excel in situations with very limited training data \cite{vu-etal-2020-exploring}, as it is the case with the labels in the ADoBo corpus \cite{adobo2021}.

In this sense, recent work has focused on systematically analyzing the performance of STILTs on both token and sequence classification. In \cite{poth2021pretrain}, the authors experimented with a diverse set of 42 intermediate and 11 target English tasks. For the token classification tasks, supplementary tasks based on POS, NER, and emergent entity recognition datasets \cite{derczynski2017results} showed to benefit from each other. However, the authors mention the risks of not choosing a proper supplementary task as a bad chosen intermeditate task can degrade performance on the target task considerably.

\section{Methods}

The corpus, released as part of the IberLEF 2021 shared task on Automatic Detection of Borrowings (ADoBO) \cite{adobo2021}, contains articles from Spanish newswire which are annotated with direct, unadapted, lexical borrowings following a set of publicly made annotation guidelines with specific assimilated borrowings, proper names and code-mixed situations.

The articles are sentence segmented and split into words. The annotations are made at the word level following the BIO annotation schema with two possible categories: \texttt{ENG} for English borrowings, and \texttt{OTHER} for lexical borrowings originating in languages other than English. The rest, including punctuation marks, are labeled using \texttt{O} and omitted from evaluation. Table \ref{table.1} shows the distributions of sentences, words, and \texttt{ENG} and \texttt{OTHER} labels per corpus split. It is specially striking the number of words labeled as \texttt{ENG} in the validation set when compared to the validation and test sets. Also worth mentioning the scarcity of words labeled as \texttt{OTHER} in general, which made learning the label solely based on this corpus a really challenging task.

\begin{table} [htbp]
\caption{\label{table.1}Distributions of sentences, words, and \texttt{ENG} and \texttt{OTHER} labels.}
\begin{center}
\begin{tabular} {lrrrr}
  \hline\rule{-4pt}{10pt}
  {\bf Split} & {\bf Sentences } & {\bf Words } & {\bf \texttt{ENG} } & {\bf \texttt{OTHER} }\\
  \hline\rule{-4pt}{10pt}
  Train & 8,216 & 231,126 & 1,701 & 32 \\
  Validation & 2,025 & 82,578 & 424 & 57 \\
  Test & 1,811 & 57,998 & 1,671 & 52 \\
\hline
\end{tabular}
\end{center}
\end{table}

The general procedure for applying STILTs follows three steps:
\begin{enumerate}
\item First, a model is trained on a semi-supervised task with no labeled data such as a language modeling task to gain some language reasoning capabilities.
\item The model is then further trained on an intermediate task for which plenty of labeled data is available.
\item Finally, the resulting model is fine-tuned further on the target task and evaluated.
\end{enumerate}

Since by nature the task involves words in more than one language, as baselines models we first fine-tuned multilingual BERT (mBERT), XLM RoBERTa (XLM-R) \cite{liu19}, a 5 languages version of mBERT including English, French, Spanish, German and Chinese (mBERT-5lang) \cite{smallermbert}, and two different monolingual Spanish BERT versions, one extracted from mBERT (mBERT-1lang) \cite{smallermbert} and the other pre-trained from scratch (BETO) \cite{canete2020spanish}.

As intermediate tasks for supplementary training, we chose mBERT models already fine-tuned on LinCE \cite{aguilar-etal-2020-lince}, a benchmark for linguistic code-switching evaluation that includes language identification (LinCE-LID), parts of speech tagging (LinCE-POS), and named entity recognition (LinCE-NER) tasks over Spanish-English code-switched data. A version of the Spanish BERT BETO fine-tuned on CoNLL-2002 for Spanish POS tags \cite{tjong-kim-sang-2002-introduction} (BETO-POS) and a version of XLM-RoBERTa fine-tuned on the same dataset for NER (XLM-R-NER) tags were also included. Distributions of sentences and words for the datasets used in these supplementary tasks are shown in Table \ref{table.2}.

\begin{table} [h]
\caption{\label{table.2}Distributions of sentences and words (sentences / words) for the datasets used in STILTs.}
\center
\setlength{\tabcolsep}{1.2mm}{
\begin{tabular} {lccc}
  \hline\rule{-4pt}{10pt}
  {\bf Dataset} & {\bf Train } & {\bf Validation } & {\bf Test } \\
  \hline\rule{-4pt}{10pt}
  CoNLL-2002 & 8,324 / 264,715 & 1,916 / 52,923 & 1,518 / 51,533 \\
  LinCE-LID & 21,030 / 253,221 & 3,332 / 40,391 & 8,289 / 97,341 \\
  LinCE-NER & 27,893 / 217,068 & 4,298 / 33,345 & 10,720 / 82,656 \\
  LinCE-POS & 33,611 / 404,428 & 10,085 / 122,656 & 23,527 / 281,579 \\
\hline
\end{tabular}
}
\end{table}

\section{Results}

We run all experiments on 2x24GB NVIDIA GPU RTX 6000, doing grid searches of hyperparameters for each model with learning rates of 1e-5, 2e-5, 3e-5, and 4e-5, and for 3, 5, and 10 epochs. We used the AdamW optimizer, with no weight decay, and a 10\% of steps for warmup. We did 3 runs with different seeds and chose the best model on the validation set while reporting results on the test set. We are reporting precision, recall, and F1 micro scores as obtained by \texttt{seqeval} \cite{seqeval18}.

\begin{table*}[h]
\caption{\label{tab2}  Best values of precision (P), recall (R), and F1 score for the validation set on baseline and STILTs models. Best scores in \textbf{bold}, second best F1 scores \underline{underlined}. }
\center
\resizebox{1\textwidth}{!}{
\setlength{\tabcolsep}{1.2mm}{
\begin{tabular}{l|ccc|ccc|ccc}
\hline
\multirow{2}{*}{\bf Model} & 
 & {\bf \texttt{ENG}} &&& {\bf \texttt{OTHER}} &&& Total \\
&
\bf P & \bf R & \bf F1 &
\bf P & \bf R & \bf F1 &
\bf P & \bf R & \bf F1 \\
\hline
mBERT & 84.47 & 88.89 & 86.62 & 65.22 & 30.61 & 41.67 & 83.19 & 80.85 & 82.00 \\ 
mBERT-1lang & 80.65 & 88.56 & 84.42 & \textbf{73.91} & 34.69 & \underline{47.22} & 80.22 & 81.13 & 80.67 \\ 
mBERT-5lang & 85.98 & \textbf{90.20} & \underline{88.04} & 62.50 & \textbf{40.82} & \textbf{49.38} & 83.85 & \textbf{83.38} & \textbf{83.62} \\ 
BETO & 79.28 & 86.27 & 82.63 & 47.62 & 20.41 & 28.57 & 77.40 & 77.18 & 77.29 \\ 
XLM-R & \textbf{86.48} & 89.87 & \textbf{88.14} & 53.85 & 28.57 & 37.33 & \textbf{84.01} & 81.41 & \underline{82.69} \\ 
\hline
LinCE-LID$_{STILT}$ & 29.29 & 18.95 & 23.02 & 12.50 & 2.04 & 3.51 & \textbf{28.64} & 16.62 & 21.03 \\ 
XLM-R-NER$_{STILT}$ & 28.46 & \textbf{23.53} & \textbf{25.76} & 50.00 & 2.04 & \underline{3.92} & 28.63 & \textbf{20.56} & \textbf{23.93} \\ 
LinCE-NER$_{STILT}$ & 23.68 & 20.59 & 22.03 & 33.33 & 2.04 & 3.85 & 23.79 & 18.03 & 20.51 \\ 
BETO-POS$_{STILT}$ & 24.43 & 17.65 & 20.49 & \textbf{100.00} & 2.04 & \textbf{4.00} & 24.77 & 15.49 & 19.06 \\ 
LinCE-POS$_{STILT}$ & \textbf{29.65} & 19.28 & \underline{23.37} & 16.67 & 2.04 & 3.64 & 29.27 & 16.9 & \underline{21.43} \\ 
\hline
\end{tabular}
}}
\end{table*}

Tables \ref{tab2} and \ref{tab3} overwhelmingly show that STILTs have absolutely no positive effect on the classification of borrowings. No matter what kind of supplementary training is used, the STILT models perform almost 4 times worse than the baseline finetuning. Among the best performing models on direct finetuning on the validation set, XLM-R and mBERT-5lang achieve somewhat similar F1 scores in total, with 83.62 F1 points for mBERT-5lang and 82.69 for XLM-R; both above mBERT. The biggest difference lies in the scores for the \texttt{OTHER} label, where all mBERT varieties perform better than XLM-R.

\begin{table*}[h]
\caption{\label{tab3} Best values of precision (P), recall (R), and F1 score for the test set on baseline and STILTs models using the best validation parameters. Best scores in \textbf{bold}, second best F1 scores \underline{underlined}. }
\center
\resizebox{1\textwidth}{!}{
\setlength{\tabcolsep}{1.2mm}{
\begin{tabular}{l|ccc|ccc|ccc}
\hline
\multirow{2}{*}{\bf Model} & 
 & {\bf \texttt{ENG}} &&& {\bf \texttt{OTHER}} &&& Total \\
&
\bf P & \bf R & \bf F1 &
\bf P & \bf R & \bf F1 &
\bf P & \bf R & \bf F1 \\
\hline
mBERT & 89.10 & 83.13 & 86.01 & 46.43 & 28.26 & \underline{35.14} & 88.09 & \textbf{81.17} & 84.49 \\ 
mBERT-1lang & 88.54 & 81.68 & 84.97 & 50.00 & 15.22 & 23.33 & 88.07 & 79.30 & 83.46 \\ 
mBERT-5lang & 90.38 & 82.65 & \underline{86.34} & 45.00 & \textbf{39.13} & \textbf{41.86} & 88.83 & 81.09 & \underline{84.78} \\
BETO & 88.18 & \textbf{83.70} & 85.88 & \textbf{66.67} & 13.04 & 21.82 & 88.02 & \textbf{81.17} & 84.45 \\ 
XLM-R & \textbf{91.47} & 82.24 & \textbf{86.61} & 43.75 & 15.22 & 22.58 & \textbf{90.8} & 79.84 & \textbf{84.97} \\ 
\hline
LinCE-LID$_{STILT}$ & \textbf{59.04} & 15.82 & \underline{24.95} & 33.33 & 6.52 & 10.91 & \textbf{58.36} & 15.49 & \underline{24.48} \\ 
XLM-R-NER$_{STILT}$ & 53.02 & \textbf{19.85} & \textbf{28.89} & 40.00 & 4.35 & 7.84 & 52.88 & \textbf{19.30} & \textbf{28.28} \\ 
LinCE-NER$_{STILT}$ & 49.47 & 15.17 & 23.22 & 37.50 & 6.52 & \underline{11.11} & 49.23 & 14.86 & 22.83 \\ 
BETO-POS$_{STILT}$ & 43.18 & 13.80 & 20.92 & \textbf{60.00} & 6.52 & \textbf{11.76} & 43.39 & 13.54 & 20.64 \\ 
LinCE-POS$_{STILT}$ & 58.54 & 15.50 & 24.51 & 30.00 & 6.52 & 10.71 & 57.69 & 15.18 & 24.03 \\ 

\hline
\end{tabular}
}}
\end{table*}

Taking the best models when evaluated on the validation test, we evaluated the performance on the test set. As shown in Table \ref{tab3} and Figure \ref{fig1}, results are similar, with mBERT-5lang performing slightly below XLM-R (84.78 vs 84.97), and presenting an F1 score for the \texttt{OTHER} label almost twice as that of the XLM-R.

\begin{figure*}
\includegraphics[width=\textwidth]{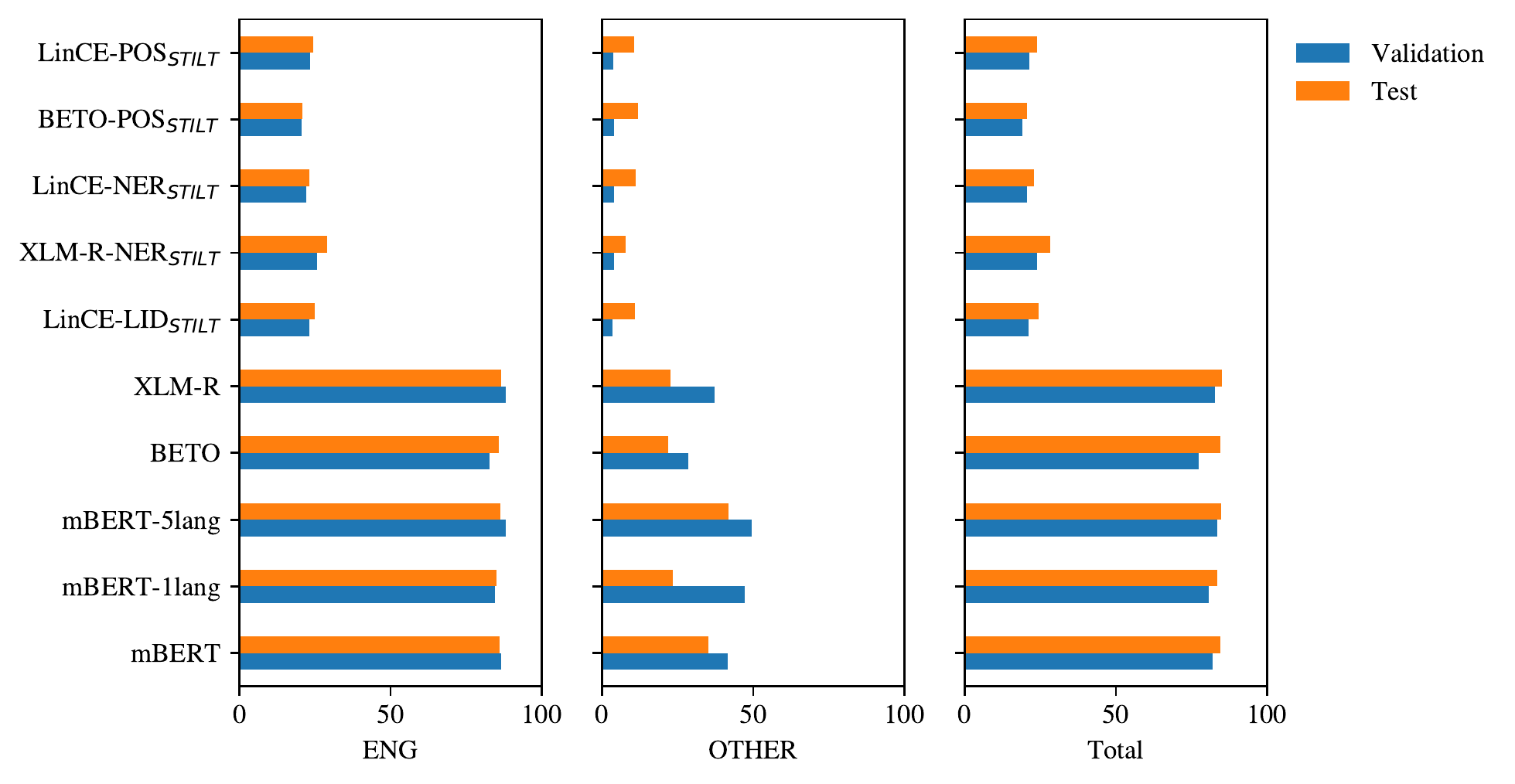}
\caption{F1 scores for the different models on the validation and test sets per label and in total.} \label{fig1}
\end{figure*}

\begin{table*}[h]
\caption{\label{tab4} Best values of precision (P), recall (R), and F1 score for the test set on baseline and STILTs models using the best test parameters. Best scores in \textbf{bold}, second best F1 scores \underline{underlined}. }
\center
\resizebox{1\textwidth}{!}{
\setlength{\tabcolsep}{1.2mm}{
\begin{tabular}{l|ccc|ccc|ccc}
\hline
\multirow{2}{*}{\bf Model} & 
 & {\bf \texttt{ENG}} &&& {\bf \texttt{OTHER}} &&& Total \\
&
\bf P & \bf R & \bf F1 &
\bf P & \bf R & \bf F1 &
\bf P & \bf R & \bf F1 \\
\hline
mBERT       & 91.10 & 84.26 & \underline{87.55} & 70.00 & 30.43 & \underline{42.42} & 90.74 & 82.33 & \underline{86.33} \\ 
mBERT-1lang & 89.84 & 81.36 & 85.39 & 54.17 & 28.26 & 37.14 & 89.09 & 79.46 & 84.00 \\ 
mBERT-5lang & 90.59 & \textbf{86.28} & \textbf{88.38} & 59.26 & \textbf{34.78} & \textbf{43.84} & 89.89 & \textbf{84.44} & \textbf{87.08} \\ 
BETO        & 90.34 & 84.58 & 87.37 & 47.06 & 17.39 & 25.40 & 89.72 & 82.18 & 85.78 \\ 
XLM-R       & \textbf{92.11} & 81.92 & 86.72 & \textbf{71.43} & 10.87 & 18.87 & \textbf{91.97} & 79.38 & 85.21 \\ 
\hline
LinCE-LID$_{STILT}$ & \textbf{55.65} & 16.71 & \underline{25.70} & \textbf{50.00} & 6.52 & \underline{11.54} & \textbf{55.56} & 16.34 & \underline{25.26} \\ 
XLM-R-NER$_{STILT}$ & 53.02 & \textbf{19.85} & \textbf{28.89} & 40.00 & 4.35 &  7.84 & 52.88 & \textbf{19.30} & \textbf{28.28} \\ 
LinCE-NER$_{STILT}$ & 53.16 & 16.95 & 25.70 & 42.86 & 6.52 & 11.32 & 52.99 & 16.58 & 25.25 \\ 
BETO-POS$_{STILT}$  & 47.47 & 13.64 & 21.19 & 37.50 & 6.52 & 11.11 & 47.25 & 13.39 & 20.86 \\ 
LinCE-POS$_{STILT}$ & \textbf{55.65} & 16.30 & 25.22 & 40.00 & \textbf{8.70} & \textbf{14.29} & 55.23 & 16.03 & 24.85 \\ 
\hline
\end{tabular}
}}
\end{table*}

Given the differences between the number of labeled words in the validation and test sets for each label, we also obtained the best scores based solely on the test set. As seen in Table \ref{tab4}, the highest F1 scores are obtained by mBERT-5lang (87.08) and mBERT (86.33), more than doubling the F1 score for the \texttt{OTHER} label with respect to the score obtained by XLM-R. In general, the models achieving higher F1 scores for the \texttt{ENG} label also get higher scores in total.

\section{Conclusions}

In this work, we framed the detection of borrowings in Spanish as a token classification task. We hypothesized that supplementary learning could improve the performance of simply fine-tuned models. However, our results strongly suggest that supplementary learning might not be as effective for token classification of borrowings as it is for sequence classification on natural language understanding tasks. This results is inline with recent research. For example, \cite{glavas-vulic-2021-supervised} noted that supervised parsing might not be as availing as expected for high-level semantic natural language understanding. Given the very nature of the ADoBo task, in which only a small set of all words tagged as verbs or nouns actually occur to be borrowings, supplementary tasks that flag all parts of speech or all named entities might have a hard time trying to figure out exactly which of those are borrowings. In this sense, although not reported as part of the results, we found that when using language identification, with no further training, LinCE-LID$_{STILT}$ correctly assigns the \texttt{ENG} label with an F1 score of 44.31. It would be interesting as future work to check how much models trained on supplementary tasks unlearn once they are trained on borrowing detection. This could also be an indication that a hybrid approach merging both language models and handcrafted features useful in language identification such as a character 3-grams could potentially boost the performance of the detection of borrowings. Additionally, adding a CRF or even a LSTM layer on top of the classifiers could see some performance gains for the simple finetuning. Auto-regressive models such as mT5 or GTP-J could also be leveraged for the task. A proper error analysis should be conducted to know exactly where and how the different approaches are failing.

We also believe that re-shuffling the train and validation sets could potentially solve the unbalance issue present in the labels of the corpus, improving the quality of the training data and making model selection more effective.

\bibliographystyle{splncs04}
\bibliography{no_stilts}

\section*{Availability}
Source code for replicating the experiments in this paper are available in a code repository: \url{https://github.com/versae/adobo-eval}. Checkpoints for the best performing models are also released as PyTorch, Tensorflow, and JAX weights:
\begin{itemize}
\item mBERT-5lang: \url{https://huggingface.co/versae/mbert-5lang-adobo2021}
\item XLM-R: \url{https://huggingface.co/versae/xlm-roberta-adobo2021}
\item mBERT-5lang (on test): \url{https://huggingface.co/versae/mbert-5lang-test-adobo2021}
\item mBERT (on test): \url{https://huggingface.co/versae/mbert-test-adobo2021}
\end{itemize}

\end{document}